\documentclass[letterpaper]{article} 
\usepackage[preprint]{aaai2027}  
\usepackage[hyphens]{url}  
\usepackage{graphicx} 
\usepackage{natbib}  
\usepackage{caption} 
\usepackage{amsmath}
\usepackage{amssymb}
\usepackage{bm}
\usepackage{multirow}
\usepackage{booktabs}

\title{HyBDM: Multi-Scale Hybrid Experts for Time Series Forecasting with Bidirectional Dependency Modeling}
\author{
    Wenqiang Ma\textsuperscript{\rm 1}\equalcontrib,
    Chen Cheng\textsuperscript{\rm 1}\equalcontrib,
    Xue Cheng\textsuperscript{\rm 1}\equalcontrib,
    Jiarui Ye\textsuperscript{\rm 1}\corresponding
}
\affiliations{
    \textsuperscript{\rm 1}Soul of Harlequin Lab\\
    
}

\begin{document}

\maketitle

\begin{abstract}
Time series forecasting (TSF) plays a vital role in many applications, yet existing models often struggle to balance the heterogeneity between long-range global patterns and short-range local variations. Some approaches capture these dependencies partially, but they often fail to exploit both temporal and feature dimensions jointly. To address this challenge, we propose HyBDM, a multi-scale hybrid model that decomposes temporal structure into global patterns and local variations, modeled by two specialized experts. The Global Patterns Expert uses an enhanced BiConv-Mamba module with bidirectional convolutions, an M-SSM layer, a forgetting mechanism, and a GDD-MLP module for cross-channel modeling. The Local Variations Expert employs a Local Window Transformer (LWT) for efficient locality-aware attention with reduced complexity. A Multi-Scale Patcher and Long-Short Router further enable multi-resolution representation and adaptive fusion of both experts. Experiments on six benchmarks show that HyBDM outperforms state-of-the-art methods in both accuracy and efficiency, demonstrating strong capability in bridging global--local dependencies for multivariate TSF.
\end{abstract}

\section{Introduction}
Time series forecasting (TSF) is a fundamental task in data analysis and machine learning, with broad implications across real-world domains. In finance, accurate TSF supports portfolio optimization and risk management \cite{fama1970efficient,tsay2005analysis}; in healthcare, it aids outbreak preparedness and resource allocation \cite{held2005statistical}; in environmental science, it helps anticipate floods and droughts \cite{shamseldin1997application}; and in wireless sensing, reconstructing spatiotemporal signal fields from sparse observations similarly requires modeling structured dependencies \cite{cheng2025mae}.

A major challenge is modeling two intertwined dependency types: long-range global patterns and short-range local variations \cite{bai2018empirical,hamilton1994time,zhang2003time}. Long-range patterns, such as seasonal temperature cycles or economic trends, reflect predictable macro-level behavior and offer useful structure for multi-step prediction \cite{hamilton1994time}. Short-range variations capture abrupt fluctuations such as demand spikes under heatwaves or sudden market shifts \cite{zhang2003time}. Effectively modeling both is essential for accurate forecasting \cite{hochreiter1997long,gu2023mamba}, while computational efficiency remains critical for real-time applications.

Traditional TSF models such as RNNs/LSTMs \cite{rumelhart1985learning,hochreiter1997long} maintain hidden states over time but struggle with long horizons due to vanishing gradients \cite{pascanu2013difficulty}. ARIMA \cite{box2015time} captures linear trends and seasonality yet handles non-linear dynamics poorly \cite{hamilton1994time}. Recent deep architectures offer stronger alternatives: CNNs detect local patterns efficiently \cite{lecun1995convolutional,bai2018empirical} but have limited receptive fields \cite{oord2016wavenet}; Transformers model global dependencies at quadratic cost \cite{vaswani2017attention,zhou2022fedformer}; SSMs such as Mamba achieve linear complexity but often mix long- and short-range effects \cite{gu2023mamba,chen2026dmamba}. Hybrid designs also face trade-offs between flexibility, accuracy, and efficiency \cite{shi2015convolutional,zhou2022fedformer}.

\begin{figure}[t]
    \centering
    \includegraphics[width=0.95\columnwidth]{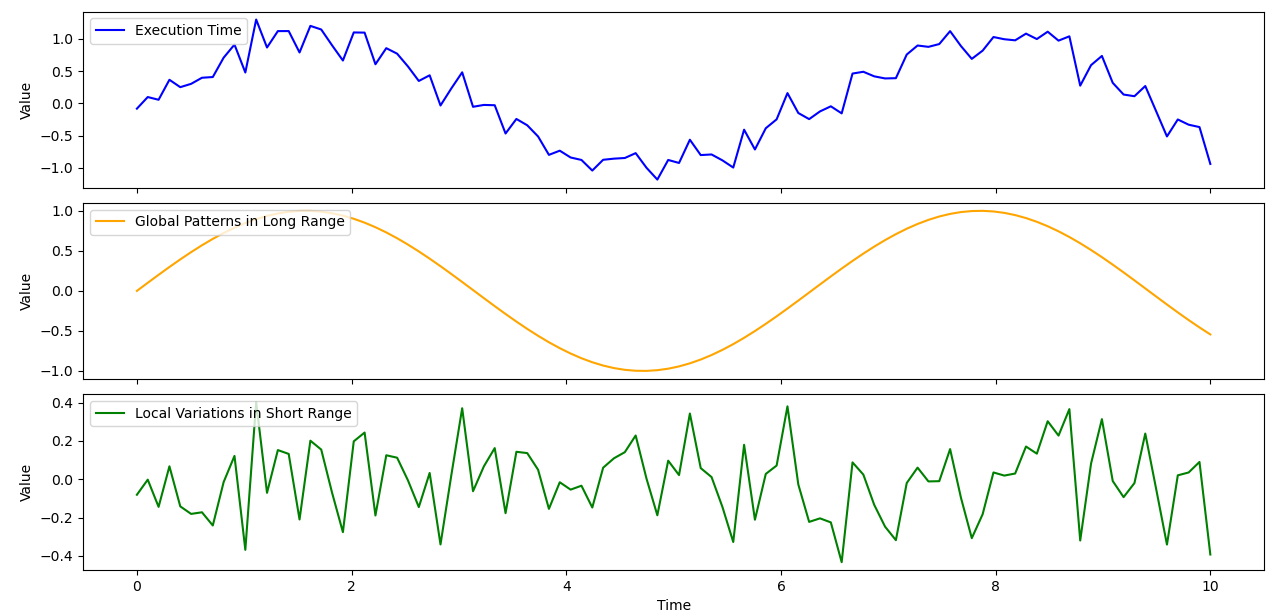}
    \caption{Decomposition of time series into global patterns (middle) and local variations (bottom). The top plot shows the original series; the middle and bottom plots highlight long-term trends and short-term fluctuations, respectively.}
    \label{fig:time_series}
\end{figure}

Existing hybrid approaches combine complementary inductive biases, but often inherit bottlenecks of their base modules---for example, CNN--RNN hybrids suffer from sequential processing overhead \cite{shi2015convolutional}---or rely on rigid decomposition schemes that adapt poorly to dynamic real-world series \cite{zhou2022fedformer}. While progress has been made, many models still struggle to balance long-range dependencies and local variations simultaneously.

To address these challenges, we present HyBDM, a multi-scale hybrid experts model with three core components. The Global Patterns Expert employs BiConv-Mamba with bidirectional convolutions, an M-SSM layer, and a forgetting mechanism to capture long-range trends and filter noise. The Local Variations Expert uses LWT with locality-aware attention for short-term fluctuations. The Long-Short Router dynamically fuses both experts based on input characteristics (Figure~\ref{fig:time_series}). Our contributions are:
\begin{itemize}
    \item We introduce HyBDM, a multi-scale hybrid framework with specialized global and local experts for long- and short-range dependency modeling.
    \item We propose BiConv-Mamba, which integrates bidirectional convolutions, an M-SSM layer, and a forgetting mechanism to model temporal dependencies in both forward and backward directions.
    \item We conduct extensive experiments on six benchmark datasets, demonstrating strong accuracy--efficiency trade-offs over state-of-the-art baselines.
\end{itemize}

\section{Related Work}
\subsection{Time Series Forecasting}
Time series forecasting has wide applications in weather prediction, finance, logistics, and foundation-model evaluation \cite{kim2026application,li2025tsfm}. Closely related spatiotemporal and world-model settings also demand multi-scale temporal reasoning, including MAE-based radio map reconstruction \cite{cheng2025mae}, physics-aware video generation \cite{cheng2026physrag}, and interactive 3D/game world models \cite{huang2026paiworld,tong2026scope}. Early RNN/LSTM models handle sequential data but struggle with long-range dependencies due to gradient issues \cite{pascanu2013difficulty}. Transformer-based models leverage self-attention to capture temporal and cross-channel dependencies \cite{Vaswani2017}. Notable variants such as Informer, Autoformer, and PatchTST \cite{Zhou2021,Wu2021,Nie2023} achieve strong performance, yet often fail to differentiate long- and short-range structure. LLM- and foundation-model-based TSF methods \cite{kim2026application,li2025tsfm} also face high computational cost and limited domain adaptation.

\subsection{State Space Models and Hybrid Architectures}
SSMs such as S4, H3, and Mamba achieve linear-time sequence modeling and strong long-context performance \cite{gu2023mamba}. Mamba uses data-dependent mechanisms to balance long- and short-term context. Hybrid SSM--Transformer models have shown strong results in language and vision \cite{Park2024,Lieber2024}, and recent TSF-oriented variants further incorporate series decomposition before SSM modeling \cite{chen2026dmamba}. Nevertheless, designs that jointly specialize global and local experts for multivariate TSF remain underexplored. Parallel trends appear in hybrid generative systems that couple retrieval with physics-aware dynamics \cite{cheng2026physrag} and in world models for robotics and interactive environments \cite{huang2026paiworld,tong2026scope}. HyBDM extends Mamba with BiConv-Mamba and a forgetting mechanism to better capture multivariate temporal dynamics.

\subsection{Cross-Channel and Global-Local Modeling}
Modeling relationships among variables is a central challenge in multivariate TSF. Two common strategies are channel-independent (CI) and channel-dependent (CD) modeling. CI treats each channel as an independent univariate series \cite{Nie2022}: it is robust under distributional shift, but ignores inter-channel coupling. CD explicitly models cross-channel interactions with attention or convolutions; it is more expressive, yet often computationally heavier and more sensitive to noise.

Despite progress on either side, efficiently combining global and local features remains difficult. Our GDD-MLP module follows the Global Patterns Expert and fuses global/local features via max/average pooling with sigmoid gating, refining representations while maintaining efficiency. HyBDM further integrates the two experts through a Long-Short Router: BiConv-Mamba captures long-range trends, LWT models short-range fluctuations, and the router dynamically balances both pathways. This multi-expert design addresses a common limitation of hybrid models that capture one dependency type well but integrate them poorly across horizons and datasets.

\section{Method}

\subsection{Problem Statement and Overview}

In multivariate time series (MTS) analysis \cite{Sen2019MTS}, each sample is a chronological sequence $\mathbf{x} = [x_1, \ldots, x_L]$ with feature vectors $x_t \in \mathbb{R}^{M}$. For example, financial series may include prices, volumes, and indices, while weather series may include temperature, humidity, and wind speed. The look-back window $L$ provides historical context, and the forecasting task is to predict $T$ future steps $[x_{L + 1}, \ldots, x_{L+T}]$.

\begin{figure*}[t]
\centering
\includegraphics[width=0.70\textwidth]{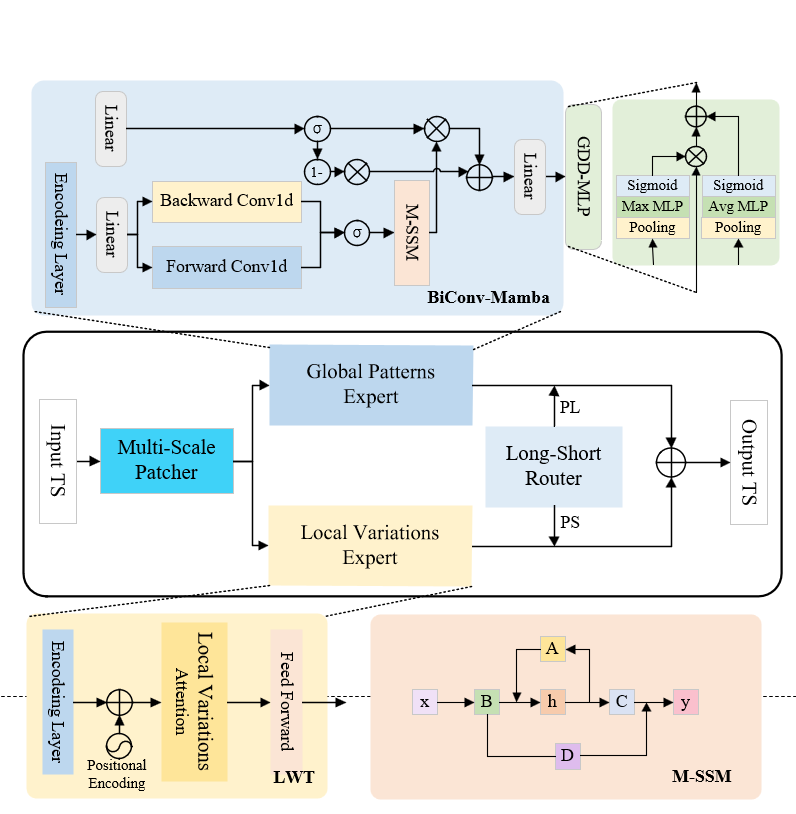}
\caption{\textbf{Overall architecture of HyBDM.} The model combines a Multi-Scale Patcher, Global Patterns Expert (BiConv-Mamba with forgetting mechanism and GDD-MLP), Local Variations Expert (LWT), and a Long-Short Router to capture both long-range and short-range dependencies.}
\label{fig:method}
\end{figure*}

Real-world MTS often exhibit hierarchical structure: macro-level trends coexist with fine-grained fluctuations across channels and horizons. HyBDM (Figure~\ref{fig:method}) targets this structure by separating long- and short-range modeling. The framework integrates SSMs and Transformer-based locality modeling through four components---BiConv-Mamba, LWT, Multi-Scale Patcher, and Long-Short Router---each addressing a distinct aspect of global--local dependency modeling \cite{Nie2023PatchTST}.

\subsection{Multi-Scale Patcher}
The Multi-Scale Patcher preprocesses the input so that global and local experts see complementary resolutions \cite{Yue2022TSMixer}. Related patch-based reconstruction ideas have also proven effective for recovering structured fields from incomplete measurements \cite{cheng2025mae}. For global patterns, larger patch lengths and strides produce low-resolution patched series that emphasize long-term trends by filtering high-frequency noise. For local variations, smaller patches preserve fine-grained fluctuations and short-range dependencies. Given multivariate series $L = (x_1, \ldots, x_L) \in \mathbb{R}^{L \times M}$, patch length $P$, and stride $Str$,
\begin{align}
N = \left\lfloor \frac{L - P}{Str} \right\rfloor + 1,
\end{align}
producing $x^{(p)} \in \mathbb{R}^{N \times P}$ and resolution $R_{\text{PTS}} = \sqrt{Str/P}$. This multi-resolution design enables adaptive scale selection and a more complete temporal representation for downstream experts.

\subsection{BiConv-Mamba Module}
BiConv-Mamba enhances long-range dependency modeling with three ingredients. Bidirectional convolutions capture temporal dynamics from both directions; an M-SSM layer enables data-dependent feature transformation; and a forgetting mechanism dynamically retains useful history while discarding noise. Together, these components improve global-pattern modeling across multiple time scales.

\subsubsection{Bidirectional Convolutions}
To capture temporal dynamics in both directions, bidirectional convolutions \cite{Bai2018TCN} extract forward/backward features $x_f, x_b$ and fuse them as
\begin{align}
x = w_f \cdot x_f + (1 - w_f) \cdot x_b,
\end{align}
where $w_f$ is learnable, ensuring both past and future temporal contexts contribute to feature extraction.

\subsubsection{M-SSM Layer}
Building upon vanilla Mamba, the M-SSM layer \cite{zeng2024c} augments SSM dynamics with data-dependent skip connections:
\begin{align}
    h_t &= A h_{t-1} + B x_t, \\
    y_t &= C h_t + D x_t,
\end{align}
where $A,B,C$ are SSM parameters and $D$ is a data-dependent skip connection derived from the input, enriching long-term retention and adaptive response to input shifts.

\subsubsection{Forgetting Mechanism}
Inspired by recurrent gating \cite{Hochreiter1997LSTM}, we compute
$\text{gate}_f = 1 - \text{sigmoid}(\text{gate}_{b2})$,
which regulates retention of historical features while integrating new information:
\begin{equation}
    \text{Output} = (\text{gate}_f \cdot x') + (\text{gate}_{b2} \cdot y),
\end{equation}
where $x'$ is the primary branch output and $y$ is the M-SSM output. This design filters outdated features and improves stability under non-stationary series.

\subsection{Local Window Transformer (LWT)}
The LWT captures short-range dependencies by limiting attention to a fixed-size sliding window of width $w$, reducing complexity from $O(S^2)$ to $O(wS)$ while preserving locality \cite{Zeng2023iTransformer}. Attention is computed only within each window, so cost grows linearly with sequence length. Stacking multiple LWT layers expands the receptive field progressively, allowing the model to gather broader context without losing local specificity. This design is particularly effective for fast-changing fluctuations in multivariate TSF, where global attention is unnecessary and often wasteful.

\subsection{Long-Short Router}
The Long-Short Router adaptively integrates outputs from BiConv-Mamba and LWT, balancing long-range and short-range experts for forecasting \cite{Shazeer2017MoE}:
\begin{align}
    p_L, p_S &= \text{softmax}(W \cdot z_R), \\
    z_{\text{LS}} &= p_L \cdot z_L + p_S \cdot z_S,
\end{align}
where $z_L$ and $z_S$ are outputs of BiConv-Mamba and LWT, respectively, and $z_R$ is the routing representation. The soft weights allow the model to emphasize global trends or local bursts depending on the input regime.

\subsection{Computational Complexity}
HyBDM maintains linear complexity $O(L)$ in sequence length \cite{Katharopoulos2020Linformer}. BiConv-Mamba combines bidirectional convolutions with SSM operations in linear time; LWT lowers attention cost from $O(S^2)$ to $O(wS)$; multi-scale patching processes multiple resolutions efficiently. Compared with quadratic Transformers, HyBDM scales favorably for long-horizon forecasting.

\section{Experiments}

\subsection{Benchmarks}

We evaluate HyBDM on six widely used multivariate benchmarks covering different domains and forecasting challenges. ETTh1 and ETTh2 contain hourly electricity transformer load from different regions, capturing distinct short-term load patterns. ETTm1 and ETTm2 provide 15-minute measurements with higher temporal resolution for finer-grained analysis. Weather includes multiple meteorological variables for multivariate climate forecasting. Traffic contains sensor-collected road flows with 862 variables, testing scalability to high-dimensional inputs. Similar long- and short-range structure also appears in logistics forecasting, such as container throughput prediction with contextual port information \cite{kim2026application}. These datasets span seasonality, granularity, and channel count, providing a comprehensive evaluation of generalization; statistics are summarized in Table~\ref{tab:datasets}.

\begin{table}[t]
    \centering
    \caption{Dataset Characteristics.}
    \label{tab:datasets}
    \small
    \setlength{\tabcolsep}{4pt}
    \begin{tabular}{lccc}
        \toprule
        \textbf{Datasets} & \textbf{Variables} & \textbf{Frequency} & \textbf{Length} \\
        \midrule
        ETTh1             & 7   & 1 hour & 17420 \\
        ETTh2             & 7   & 1 hour & 17420 \\
        ETTm1             & 7   & 15 min & 69680 \\
        ETTm2             & 7   & 15 min & 69680 \\
        Weather           & 21  & 10 min & 52696 \\
        Traffic           & 862 & 1 hour & 17544 \\
        \bottomrule
    \end{tabular}
\end{table}

\subsection{Experiment Setup}

All experiments use PyTorch on an NVIDIA RTX 4090 GPU with the ADAM optimizer and $L_2$ loss. We follow the standard train/validation/test splits for each benchmark and evaluate all models under identical look-back windows and prediction horizons. Batch size is chosen per dataset following common protocols in prior work, and all models are trained for 20 epochs with early stopping on validation MSE. Hyperparameters for HyBDM and baselines are tuned on validation sets to ensure fair comparison.

\subsection{Evaluation Metrics}

We report MSE and MAE (lower is better). MSE emphasizes large deviations via squared error, while MAE is more robust to outliers. Both metrics are averaged over all prediction steps and channels, following standard practice in multivariate TSF benchmarks:
\begin{align}
MSE = \frac{1}{n}\sum_{i = 1}^{n}(y_i - \hat{y}_i)^2,\quad
MAE=\frac{1}{n}\sum_{i = 1}^{n}\vert y_i-\hat{y}_i\vert ,
\end{align}
where $n$ is the number of test samples, $y_i$ is the ground truth, and $\hat{y}_i$ is the model prediction.

\subsection{Comparison with State-of-the-Art Methods}

We compare HyBDM against ten recent baselines across forecast horizons $\{96,192,336,720\}$: SST, CMamba, iTransformer, RLinear, Crossformer, TimesNet, DLinear, TimeMachine, PatchTST, and Tide. Among them, SST is a closely related hybrid Mamba--Transformer expert model \cite{xu2024sst}, while recent foundation-model benchmarks further highlight the need for unified TSF evaluation \cite{li2025tsfm}. PatchTST uses patch-structured attention for long-range modeling; TimeMachine captures bidirectional temporal dependencies; TimesNet analyzes cyclic patterns in the time-frequency domain; CMamba enhances cross-variable interactions; Crossformer models temporal--variable dependencies via cross-attention; iTransformer focuses on inverted temporal representation; RLinear, DLinear, and Tide provide strong linear and decomposition-based references. Full MSE/MAE results are reported in Tables~\ref{tab:forecast_comparison} and~\ref{tab:forecast_comparison_cont}.

\begin{table*}[t]
\centering
    \caption{Forecasting results (Part 1). H: forecast horizon.}
    \label{tab:forecast_comparison}
    \footnotesize
    \setlength{\tabcolsep}{2pt}
    \begin{tabular}{@{}l c cc cc cc cc cc cc@{}}
  \toprule
    \multirow{2}{*}{Dataset} & \multirow{2}{*}{H}
    & \multicolumn{2}{c}{HyBDM} & \multicolumn{2}{c}{SST} & \multicolumn{2}{c}{CMamba}
    & \multicolumn{2}{c}{iTrans.} & \multicolumn{2}{c}{RLinear} & \multicolumn{2}{c}{Cross.} \\
    \cmidrule(lr){3-4} \cmidrule(lr){5-6} \cmidrule(lr){7-8} \cmidrule(lr){9-10} \cmidrule(lr){11-12} \cmidrule(lr){13-14}
    & & MSE & MAE & MSE & MAE & MSE & MAE & MSE & MAE & MSE & MAE & MSE & MAE \\
    \midrule
    \multirow{5}{*}{ETTh1}
    & 96  & \textbf{0.375} & 0.401 & 0.391 & 0.416 & \textbf{0.375} & \textbf{0.388} & 0.386 & 0.405 & 0.386 & \underline{0.395} & 0.423 & 0.448 \\
    & 192 & \textbf{0.411} & 0.432 & \underline{0.421} & 0.426 & 0.428 & \textbf{0.417} & 0.441 & 0.436 & 0.439 & \underline{0.424} & 0.471 & 0.474 \\
    & 336 & \textbf{0.408} & \textbf{0.434} & \underline{0.439} & 0.443 & 0.465 & \underline{0.437} & 0.487 & 0.458 & 0.479 & 0.446 & 0.570 & 0.546 \\
    & 720 & \textbf{0.452} & 0.471 & 0.508 & 0.508 & \underline{0.470} & \textbf{0.458} & 0.503 & 0.491 & 0.481 & 0.470 & 0.653 & 0.621 \\
    & Avg & \textbf{0.412} & 0.435 & 0.440 & 0.448 & \underline{0.435} & \textbf{0.425} & 0.454 & 0.447 & 0.446 & \underline{0.434} & 0.529 & 0.522 \\
    \midrule
    \multirow{5}{*}{ETTh2}
    & 96  & \underline{0.278} & 0.343 & 0.289 & 0.344 & 0.280 & \textbf{0.329} & 0.297 & 0.349 & 0.288 & \underline{0.338} & 0.745 & 0.584 \\
    & 192 & \textbf{0.337} & \textbf{0.375} & 0.375 & 0.401 & \underline{0.362} & \underline{0.381} & 0.380 & 0.400 & 0.374 & 0.390 & 0.877 & 0.656 \\
    & 336 & \textbf{0.334} & \textbf{0.386} & \underline{0.374} & \underline{0.413} & 0.417 & 0.422 & 0.428 & 0.432 & 0.415 & 0.426 & 1.043 & 0.732 \\
    & 720 & \textbf{0.384} & \textbf{0.424} & 0.418 & 0.446 & \underline{0.417} & \textbf{0.436} & 0.427 & 0.445 & 0.420 & 0.440 & 1.104 & 0.763 \\
    & Avg & \textbf{0.333} & \textbf{0.382} & \underline{0.364} & 0.401 & 0.369 & \underline{0.392} & 0.383 & 0.407 & 0.374 & 0.398 & 0.942 & 0.684 \\
    \midrule
    \multirow{5}{*}{ETTm1}
    & 96  & \textbf{0.301} & \underline{0.353} & \textbf{0.301} & \underline{0.353} & \underline{0.309} & \textbf{0.337} & 0.334 & 0.368 & 0.355 & 0.376 & 0.406 & 0.426 \\
    & 192 & \textbf{0.344} & \underline{0.380} & \textbf{0.344} & \underline{0.380} & \underline{0.360} & \textbf{0.366} & 0.377 & 0.391 & 0.391 & 0.392 & 0.450 & 0.451 \\
    & 336 & \textbf{0.369} & \underline{0.395} & \textbf{0.369} & \underline{0.395} & \underline{0.394} & \textbf{0.389} & 0.426 & 0.420 & 0.424 & 0.415 & 0.532 & 0.515 \\
    & 720 & \textbf{0.425} & \textbf{0.425} & \underline{0.426} & \underline{0.428} & 0.465 & \underline{0.428} & 0.491 & 0.459 & 0.487 & 0.450 & 0.666 & 0.589 \\
    & Avg & \textbf{0.360} & \underline{0.389} & \textbf{0.360} & \underline{0.389} & \underline{0.382} & \textbf{0.380} & 0.407 & 0.410 & 0.414 & 0.407 & 0.513 & 0.496 \\
    \midrule
    \multirow{5}{*}{ETTm2}
    & 96  & \textbf{0.168} & 0.262 & 0.175 & 0.264 & \underline{0.171} & \textbf{0.248} & 0.180 & 0.264 & 0.182 & 0.265 & 0.287 & 0.366 \\
    & 192 & \textbf{0.234} & 0.307 & \underline{0.235} & 0.307 & \textbf{0.234} & \textbf{0.291} & 0.250 & 0.309 & 0.246 & 0.304 & 0.414 & 0.492 \\
    & 336 & \textbf{0.287} & \textbf{0.339} & \underline{0.293} & \textbf{0.339} & 0.302 & \textbf{0.339} & 0.311 & 0.348 & 0.307 & \underline{0.342} & 0.597 & 0.542 \\
    & 720 & \textbf{0.365} & 0.391 & \underline{0.366} & \textbf{0.385} & 0.389 & \underline{0.390} & 0.412 & 0.407 & 0.407 & 0.398 & 1.730 & 1.402 \\
    & Avg & \textbf{0.264} & 0.332 & \underline{0.267} & \underline{0.323} & 0.274 & \textbf{0.317} & 0.288 & 0.410 & 0.286 & 0.327 & 0.757 & 0.610 \\
    \midrule
    \multirow{5}{*}{Weather}
    & 96  & \textbf{0.147} & \underline{0.197} & 0.153 & 0.205 & \underline{0.150} & \textbf{0.187} & 0.174 & 0.214 & 0.192 & 0.232 & 0.158 & 0.230 \\
    & 192 & \textbf{0.192} & \underline{0.240} & \underline{0.196} & 0.244 & 0.200 & \textbf{0.236} & 0.221 & 0.254 & 0.240 & 0.271 & 0.206 & 0.277 \\
    & 336 & \textbf{0.242} & \underline{0.281} & \underline{0.246} & 0.284 & 0.260 & \underline{0.281} & 0.278 & \textbf{0.269} & 0.292 & 0.307 & 0.272 & 0.335 \\
    & 720 & \textbf{0.309} & \textbf{0.329} & \underline{0.314} & \underline{0.334} & 0.347 & 0.339 & 0.358 & 0.349 & 0.364 & 0.353 & 0.398 & 0.418 \\
    & Avg & \textbf{0.223} & \underline{0.262} & \underline{0.227} & 0.267 & 0.239 & \textbf{0.261} & 0.258 & 0.279 & 0.272 & 0.291 & 0.259 & 0.315 \\
    \midrule
    \multirow{5}{*}{Traffic}
    & 96  & \textbf{0.365} & \underline{0.255} & \underline{0.367} & 0.257 & 0.414 & \textbf{0.251} & 0.395 & 0.268 & 0.649 & 0.389 & 0.522 & 0.290 \\
    & 192 & \textbf{0.380} & 0.270 & \underline{0.385} & \underline{0.266} & 0.432 & \textbf{0.257} & 0.417 & 0.276 & 0.601 & 0.366 & 0.530 & 0.293 \\
    & 336 & 0.413 & 0.289 & \textbf{0.401} & \underline{0.275} & 0.446 & \textbf{0.265} & 0.433 & 0.283 & 0.609 & 0.369 & 0.558 & 0.305 \\
    & 720 & \textbf{0.443} & \textbf{0.285} & \underline{0.445} & 0.302 & 0.485 & \underline{0.286} & 0.467 & 0.302 & 0.647 & 0.387 & 0.589 & 0.328 \\
    & Avg & \textbf{0.400} & \underline{0.275} & \textbf{0.400} & \underline{0.275} & 0.444 & \textbf{0.265} & 0.428 & 0.282 & 0.620 & 0.378 & 0.550 & 0.304 \\
    \midrule
    \textbf{1st Count} & & \textbf{28} & \underline{9} & \underline{6} & 2 & 2 & \textbf{21} & 0 & 1 & 0 & 0 & 0 & 0 \\
  \bottomrule
\end{tabular}
\end{table*}

\begin{table*}[t]
\centering
    \caption{Forecasting results (Part 2). TNet: TimesNet; DLin: DLinear; P.TST: PatchTST. H: forecast horizon.}
    \label{tab:forecast_comparison_cont}
    \small
    \setlength{\tabcolsep}{3pt}
    \begin{tabular}{@{}l c cc cc cc cc cc@{}}
  \toprule
    \multirow{2}{*}{Dataset} & \multirow{2}{*}{H}
    & \multicolumn{2}{c}{TNet} & \multicolumn{2}{c}{DLin} & \multicolumn{2}{c}{TMach.}
    & \multicolumn{2}{c}{P.TST} & \multicolumn{2}{c}{Tide} \\
    \cmidrule(lr){3-4} \cmidrule(lr){5-6} \cmidrule(lr){7-8} \cmidrule(lr){9-10} \cmidrule(lr){11-12}
    & & MSE & MAE & MSE & MAE & MSE & MAE & MSE & MAE & MSE & MAE \\
    \midrule
    \multirow{5}{*}{ETTh1}
    & 96  & \underline{0.384} & 0.402 & 0.386 & 0.400 & 0.398 & 0.402 & 0.414 & 0.419 & 0.479 & 0.464 \\
    & 192 & 0.436 & 0.429 & 0.437 & 0.432 & 0.435 & 0.440 & 0.460 & 0.445 & 0.525 & 0.492 \\
    & 336 & 0.491 & 0.469 & 0.481 & 0.459 & 0.450 & 0.448 & 0.501 & 0.466 & 0.565 & 0.515 \\
    & 720 & 0.521 & 0.500 & 0.519 & 0.516 & 0.480 & \underline{0.465} & 0.500 & 0.488 & 0.594 & 0.558 \\
    & Avg & 0.485 & 0.450 & 0.456 & 0.452 & 0.441 & 0.439 & 0.469 & 0.455 & 0.541 & 0.507 \\
    \midrule
    \multirow{5}{*}{ETTh2}
    & 96  & 0.340 & 0.374 & 0.333 & 0.387 & \textbf{0.230} & 0.349 & 0.302 & 0.348 & 0.400 & 0.440 \\
    & 192 & 0.402 & 0.414 & 0.477 & 0.476 & 0.371 & 0.400 & 0.388 & 0.400 & 0.528 & 0.509 \\
    & 336 & 0.452 & 0.452 & 0.594 & 0.541 & 0.402 & 0.449 & 0.426 & 0.433 & 0.643 & 0.571 \\
    & 720 & 0.462 & 0.468 & 0.831 & 0.657 & 0.425 & \underline{0.438} & 0.431 & 0.446 & 0.847 & 0.679 \\
    & Avg & 0.414 & 0.427 & 0.559 & 0.515 & 0.357 & 0.409 & 0.387 & 0.407 & 0.611 & 0.550 \\
    \midrule
    \multirow{5}{*}{ETTm1}
    & 96  & 0.338 & 0.375 & 0.345 & 0.372 & 0.312 & 0.371 & 0.329 & 0.367 & 0.364 & 0.387 \\
    & 192 & 0.374 & 0.387 & 0.380 & 0.389 & 0.365 & 0.409 & 0.367 & 0.385 & 0.398 & 0.404 \\
    & 336 & 0.410 & 0.411 & 0.413 & 0.413 & 0.421 & 0.410 & 0.399 & 0.410 & 0.428 & 0.425 \\
    & 720 & 0.478 & 0.450 & 0.474 & 0.453 & 0.496 & 0.437 & 0.454 & 0.439 & 0.487 & 0.461 \\
    & Avg & 0.400 & 0.406 & 0.403 & 0.407 & 0.399 & 0.407 & 0.387 & 0.400 & 0.419 & 0.419 \\
    \midrule
    \multirow{5}{*}{ETTm2}
    & 96  & 0.187 & 0.267 & 0.193 & 0.292 & 0.185 & 0.290 & 0.175 & \underline{0.259} & 0.207 & 0.305 \\
    & 192 & 0.249 & 0.309 & 0.284 & 0.362 & 0.292 & 0.309 & 0.241 & \underline{0.302} & 0.290 & 0.364 \\
    & 336 & 0.321 & 0.351 & 0.369 & 0.427 & 0.321 & 0.367 & 0.305 & 0.343 & 0.377 & 0.422 \\
    & 720 & 0.408 & 0.403 & 0.554 & 0.522 & 0.401 & 0.400 & 0.402 & 0.400 & 0.558 & 0.524 \\
    & Avg & 0.291 & 0.333 & 0.350 & 0.401 & 0.300 & 0.342 & 0.281 & 0.326 & 0.358 & 0.404 \\
    \midrule
    \multirow{5}{*}{Weather}
    & 96  & 0.172 & 0.220 & 0.196 & 0.255 & 0.174 & 0.218 & 0.177 & 0.218 & 0.202 & 0.261 \\
    & 192 & 0.219 & 0.261 & 0.237 & 0.296 & 0.200 & 0.258 & 0.225 & 0.259 & 0.242 & 0.298 \\
    & 336 & 0.280 & 0.306 & 0.283 & 0.335 & 0.280 & 0.299 & 0.278 & 0.297 & 0.287 & 0.335 \\
    & 720 & 0.365 & 0.359 & 0.345 & 0.381 & 0.352 & 0.359 & 0.354 & 0.348 & 0.351 & 0.386 \\
    & Avg & 0.259 & 0.287 & 0.265 & 0.317 & 0.252 & 0.284 & 0.259 & 0.280 & 0.271 & 0.320 \\
    \midrule
    \multirow{5}{*}{Traffic}
    & 96  & 0.593 & 0.321 & 0.650 & 0.396 & 0.398 & 0.274 & 0.462 & 0.295 & 0.805 & 0.493 \\
    & 192 & 0.617 & 0.336 & 0.598 & 0.370 & 0.393 & 0.282 & 0.366 & 0.296 & 0.756 & 0.474 \\
    & 336 & 0.629 & 0.336 & 0.605 & 0.373 & 0.443 & 0.368 & 0.482 & 0.304 & 0.760 & 0.473 \\
    & 720 & 0.640 & 0.350 & 0.645 & 0.394 & 0.470 & 0.309 & 0.514 & 0.322 & 0.719 & 0.449 \\
    & Avg & 0.620 & 0.336 & 0.625 & 0.383 & \underline{0.426} & 0.308 & 0.456 & 0.304 & 0.760 & 0.473 \\
    \midrule
    \textbf{1st Count} & & 0 & 0 & 0 & 0 & 1 & 0 & 0 & 0 & 0 & 0 \\
  \bottomrule
\end{tabular}
\end{table*}

HyBDM ranks first in most settings and remains stable across datasets. On ETTh1/ETTm1, bidirectional modeling and multi-scale patching capture long-range seasonal structure, while several baselines degrade as the horizon grows from 96 to 720. Table~\ref{tab:forecast_comparison} shows 28/30 MSE wins against SST, CMamba, iTransformer, RLinear, and Crossformer; Crossformer degrades sharply on ETTh2. On Weather, the forgetting gate and local expert handle rapid fluctuations; on Traffic, HyBDM scales to 862 channels while remaining competitive with SST.

Compared with Part~2 baselines in Table~\ref{tab:forecast_comparison_cont}, HyBDM remains the most consistent overall. TimesNet and PatchTST are strong on selected ETT horizons but less stable across all six datasets. TimeMachine achieves the best ETTh2 MSE at horizon 96, yet its average performance lags behind HyBDM on most other settings. Linear baselines (RLinear, DLinear, Tide) are efficient but struggle when long- and short-range effects interact strongly, which is precisely the regime targeted by our hybrid design. These trends indicate that explicitly separating global and local pathways is more reliable than relying on a single inductive bias across heterogeneous benchmarks.

\subsection{Baseline Comparison}
We further compare HyBDM with three widely used baselines---RLinear, DLinear, and Tide---on complex datasets such as Weather and Traffic. RLinear applies simple linear fitting for prediction; its basic operations ensure high efficiency but limit non-linear modeling. DLinear extracts features through differencing, which works well under limited resources yet can over-emphasize local changes on long-cycle series. Tide relies on decomposition-based forecasting and often misses cross-channel dependencies in multivariate settings.

According to Tables~\ref{tab:forecast_comparison} and~\ref{tab:forecast_comparison_cont}, HyBDM outperforms these baselines consistently. On Weather, the global expert captures longer-range meteorological correlations, while the local expert responds to rapid regime shifts. On Traffic, HyBDM integrates flow dynamics across 862 channels and better reflects short-term rush-hour surges as well as longer weekday--weekend patterns. TimesNet and PatchTST remain competitive on selected ETT horizons, while TimeMachine is strongest on ETTh2 MSE at horizon 96; Tide remains weak on Traffic. Overall, HyBDM offers the most balanced accuracy--efficiency trade-off among all compared methods.

\subsection{Ablation Study}
\label{subsec:ablation_study}

To evaluate the contributions of key components in HyBDM, we conduct an ablation study focusing on the BiConv-Mamba module and the forgetting mechanism. We compare four variants: (i)~HyBDM (full), (ii)~w/o BiConv, (iii)~w/o Forget Gate, and (iv)~SST. Removing BiConv-Mamba disables bidirectional global modeling, while removing the forgetting gate tests whether adaptive history filtering is necessary for non-stationary series. SST serves as a closely related hybrid baseline under the same training protocol.

Table~\ref{tab:ablation_combined} reports average MSE/MAE across horizons. Removing BiConv-Mamba increases errors on datasets with strong long-range structure (ETTh1, ETTm1, and Traffic), confirming its role in global pattern extraction. Removing the forgetting gate mainly hurts Weather, where short-term fluctuations dominate. The two components are complementary rather than redundant: BiConv-Mamba stabilizes trend modeling, while the forgetting gate suppresses outdated local context. Full HyBDM outperforms SST on most datasets, validating the benefit of bidirectional SSM modeling plus adaptive gating beyond a standard hybrid design.

\begin{table}[!ht]
\centering
\caption{Ablation results (MSE/MAE).}
\label{tab:ablation_combined}
\small
\setlength{\tabcolsep}{3pt}
\begin{tabular}{l|cc|cc|cc|cc}
\toprule
\multirow{2}{*}{\textbf{Dataset}} & \multicolumn{2}{c|}{\textbf{HyBDM}} & \multicolumn{2}{c|}{\textbf{w/o BiConv}} & \multicolumn{2}{c|}{\textbf{w/o Forget}} & \multicolumn{2}{c}{\textbf{SST}} \\
\cmidrule(lr){2-3} \cmidrule(lr){4-5} \cmidrule(lr){6-7} \cmidrule(lr){8-9}
& MSE & MAE & MSE & MAE & MSE & MAE & MSE & MAE \\
\midrule
ETTh1   & \textbf{0.412} & \textbf{0.435} & 0.413 & 0.442 & 0.431 & 0.444 & 0.440 & 0.448 \\
ETTh2   & \textbf{0.333} & \textbf{0.382} & 0.340 & 0.388 & 0.342 & 0.388 & 0.364 & 0.401 \\
ETTm1   & \textbf{0.360} & \textbf{0.389} & 0.383 & 0.406 & 0.384 & 0.406 & 0.360 & 0.389 \\
ETTm2   & \textbf{0.264} & 0.332         & 0.291 & 0.338 & 0.275 & 0.329 & 0.267 & \textbf{0.323} \\
Weather & \textbf{0.223} & \textbf{0.262} & 0.231 & 0.268 & 0.228 & 0.265 & 0.227 & 0.267 \\
\bottomrule
\end{tabular}
\end{table}

The ablation results highlight module-level specialization. BiConv-Mamba removal has limited impact on ETTh2 and ETTm2, where local fluctuations are more prominent, but causes clear degradation on ETTh1 and ETTm1. The forgetting gate contributes most on Weather, where rapid temperature and humidity changes require selective retention of recent observations. Together, these findings show that HyBDM's gains stem from jointly modeling global trends and local bursts rather than from any single block in isolation.

\subsection{Computational Complexity Analysis}
To further analyze efficiency, we evaluate HyBDM in terms of both time and memory consumption.

\paragraph{Time.}
HyBDM maintains linear time complexity $O(L)$ via BiConv-Mamba and LWT ($O(wS)$ instead of $O(S^2)$). Multi-scale patching further reduces redundant computation across resolutions, unlike quadratic Transformers whose self-attention cost grows rapidly with look-back length. As a result, HyBDM remains practical for long-horizon forecasting on high-dimensional inputs such as Traffic.

\paragraph{Memory.}
Patching and routing slightly increase memory, but preserve coarse trends and fine fluctuations while adapting expert fusion. The multi-scale patcher stores representations at complementary resolutions, and the Long-Short Router maintains both expert outputs before fusion. Although these modules add overhead, they provide a stronger complexity--accuracy trade-off than PatchTST and SST in our setting, especially when both seasonal structure and abrupt local changes must be captured simultaneously.

\section{Conclusion}
We present HyBDM, a multi-scale hybrid experts framework for time series forecasting. By combining BiConv-Mamba, LWT, Multi-Scale Patching, and a Long-Short Router, HyBDM captures long-range global patterns and short-range local variations with linear complexity $O(L)$. Experiments on six benchmarks and ablations confirm consistent gains from bidirectional global modeling and adaptive forgetting. These results suggest that explicitly routing between global and local experts is an effective strategy for multivariate TSF in energy, weather, and traffic applications.

\bibliography{sample}

\end{document}